\documentclass{svproc}
\usepackage{url}

\usepackage{graphicx}
\usepackage{cite}

\usepackage{xcolor}

\usepackage{caption}
\usepackage{subfigure}
\usepackage{algorithm}
\usepackage{algpseudocode}
\usepackage{amsmath}
\usepackage{amssymb}
\usepackage{pgfplots}
\usepackage{commath}

\definecolor{Yellow}{rgb}{1,0.9,0.7}
\definecolor{Pink}{rgb}{1,0.85,0.85}
\definecolor{AntiqueWhite}{rgb}{0.9,0.9,0.9}

\newcommand{\NOTE}[1]%
{
\noindent
\fboxsep=2mm\fcolorbox{black}{AntiqueWhite}{\parbox{0.95\columnwidth}
{\textbf{NOTE: } #1}
}
}

\algnewcommand\algorithmicswitch{\textbf{switch}}
\algnewcommand\algorithmiccase{\textbf{case}}
\algnewcommand\algorithmicassert{\texttt{assert}}
\algnewcommand\Assert[1]{\State \algorithmicassert(#1)}%
\algdef{SE}[SWITCH]{Switch}{EndSwitch}[1]{\algorithmicswitch\ #1\ \algorithmicdo}{\algorithmicend\ \algorithmicswitch}%
\algdef{SE}[CASE]{Case}{EndCase}[1]{\algorithmiccase\ #1}{\algorithmicend\ \algorithmiccase}%
\algtext*{EndCase}%

\begin{document}
\mainmatter              
\title{Dynamic Formation Reshaping Based on Point Set Registration in a Swarm of Drones
\thanks{This work has been supported by the Academy of Finland-funded research project 314048.}}
\titlerunning{Dynamic Formation Reshaping Based on PSR in a Swarm of Drones}  
%

\author{Jawad N. Yasin\inst{1} \and Sherif A.S. Mohamed\inst{1}
Mohammad-Hashem Haghbayan\inst{1} \and Jukka Heikkonen\inst{1} \and Hannu Tenhunen\inst{2} \and Muhammad Mehboob Yasin\inst{3} \and
Juha Plosila\inst{1}}
\authorrunning{Jawad N. Yasin et al.}

%
%
\institute{Autonomous Systems Laboratory, Department of Future Technologies, University of Turku, Vesilinnantie 5, 20500 Turku, Finland\\
\email{\{janaya, samoha, mohhag, jukhei, juplos\}@utu.fi},\\
\and
Department of Industrial and Medical Electronics, KTH Royal Institute of Technology, Brinellvägen 8, 114 28 Stockholm, Sweden\\ \email{hannu@kth.se}
\and
Department of Computer Networks, College of Computer Sciences \& Information Technology, King Faisal University, Hofuf, Saudi Arabia\\ \email{mmyasin@kfu.edu.sa}}

\maketitle              

\begin{abstract}
This work focuses on the formation reshaping in an optimized manner in autonomous swarm of drones. Here, the two main problems are: 1) how to break and reshape the initial formation in an optimal manner, and 2) how to do such reformation while minimizing the overall deviation of the drones and the overall time, i.e., without slowing down. To address the first problem, we introduce a set of routines for the drones/agents to follow while reshaping to a secondary formation shape. And the second problem is resolved by utilizing the temperature function reduction technique, originally used in the point set registration process. The goal is to be able to dynamically reform the shape of multi-agent based swarm in near-optimal manner while going through narrow openings between, for instance obstacles, and then bringing the agents back to their original shape after passing through the narrow passage using point set registration technique.

\end{abstract}
\section{Introduction}
Unmanned aerial vehicles (UAVs) are gaining more attention of the researchers due to their numerous advantages, such as robustness, agility, cost-effectiveness, over ground vehicles. These benefits have created even more demand for the autonomous UAVs and especially their swarms in various applications and fields, for instance threat detection, intelligent transportation systems, search and rescue, military purposes, surveying, and mapping \cite{ladd2009non, 8682048, HE2018327}.

Drones and swarms of drones, despite getting more sophisticated, still face several mission or design limitations and challenges, such as optimal navigation and obstacle avoidance, dynamic reformation, payload, flight time due to limited battery life, design of state-of-the-art stability controllers, and optimal resource allocation \cite{8500274, tseng2017flight, jdiscsic}. UAVs can encounter stationary or dynamic obstacles while navigating and therefore require a reliable collision avoidance system on-board for safe navigation \cite{7947166, 776ccee71a23417a93377ba6abc18a23}. Different types of collision avoidance methods can be generalized into four categories \cite{jdsurvey, 10.1007/978-3-030-37393-1_32}: potential field based methods \cite{doi:10.2514/6.2020-0487, SENANAYAKE2016422}; geometric methods \cite{6722091, 8682048, 10.1007/978-3-030-37393-1_32}; sense and avoid methods \cite{prats2012requirements, jdiscsic, 5412074}; optimization based methods \cite{zhang2017optimization, doi:10.2514/6.2014-0966}. Furthermore, the methodologies for formation maintenance can be categorized into three approaches \cite{1657384, OH2015424}: leader follower based \cite{Shen2014, jdpaams}; virtual structure based \cite{li2008formation, DONG2016415}; behaviour based \cite{lawton2003decentralized, 736776}.

One of the biggest limitation for UAVs is the mission life due to their limited battery capacity \cite{tseng2017flight, jdpaams}. More deviation from the path would result in more power consumption leading to shortened mission life. Similarly, if the swarm slows down or hovers at a certain location for reshaping before continuing, it increases the mission time meaning more power consumed and shortened mission life on a single charge. Maintaining formation and only reshaping the formation while continuing towards the destination can lower the unexpected deviations and power consumption or losing a drone, as while navigating through the obstacles (e.g. going through the window of a building) breaking the formation to bypass the obstacles in different directions may cause inter-drone communication failure or excessive deviation, resulting in more power consumption.

In this paper, for maintaining the formation, we utilize the leader-follower based approach due to its simplicity, non-complex implementation, reliability, and scalability \cite{jdiscsic}. Keeping the above mentioned limitations and time criticality, we propose our algorithm where a swarm formation comes across multiple obstacles and can go through the gap between them while dynamically changing the formation and reforming to original shape without slowing down. In the proposed algorithm, upon detection of the obstacles and the available gap between them, the swarm reshapes into a queue formation. In order to do so, the agents in the swarm select their temporary leaders and navigate to reach and maintain the minimum defined distance between two agents. Once the agents come out of the obstacles, to reform the original formation shape, they immediately start navigating to the best possible position determined through the point-set registration technique.

The rest of the paper is organized as follows. Section 2 provides the development of the proposed approach. Section 3 is based on simulation results. And finally, conclusions, discussions, and future work is provided in Section 4.

\section{Proposed Approach}

\begin{figure}[h]
\begin{center}
    \subfigure[\label{fig:obsang}]{\includegraphics[width=0.485\linewidth]{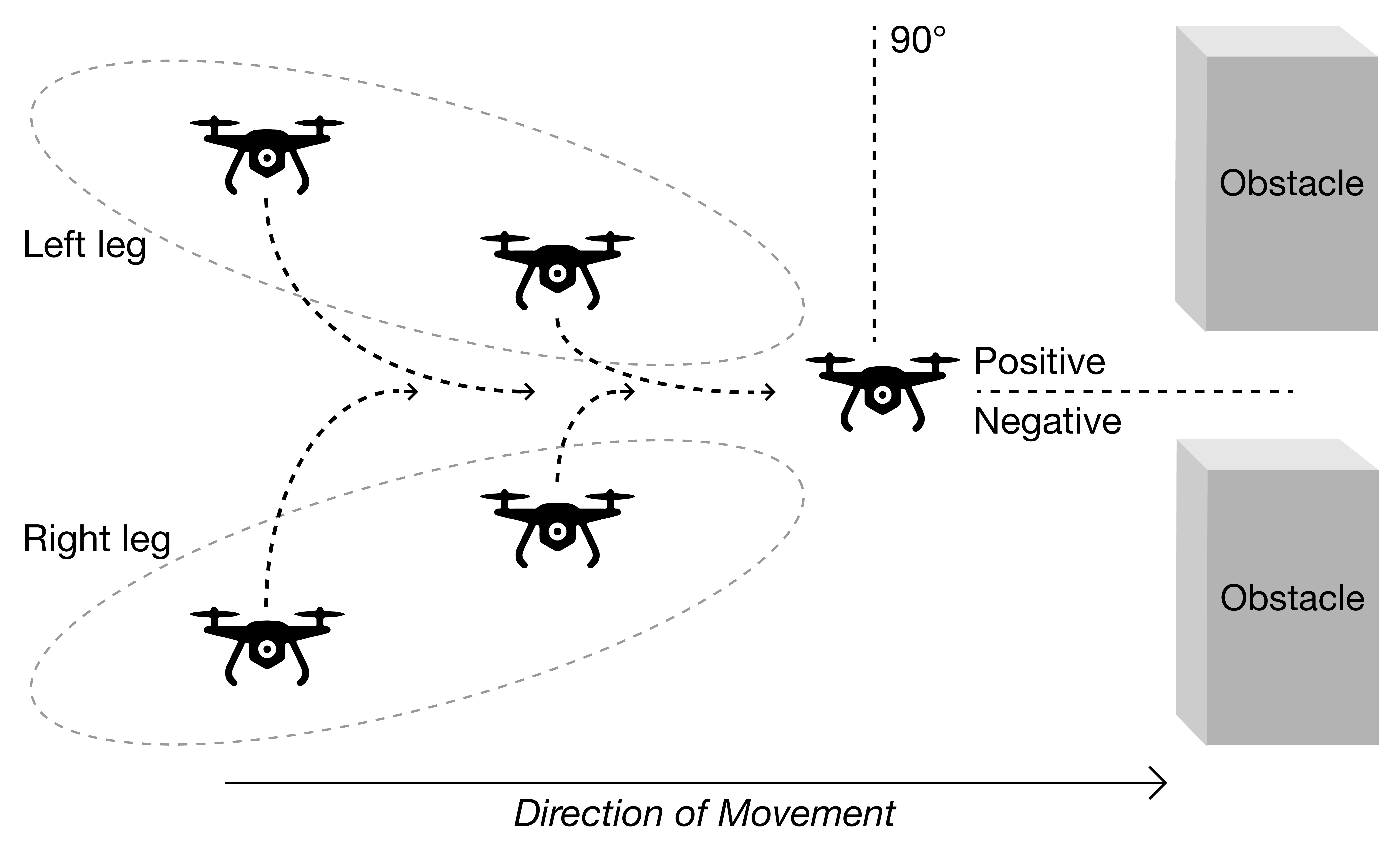}}\hspace{0.2cm}
    \subfigure[\label{fig:reshape}]{\includegraphics[width=0.485\linewidth]{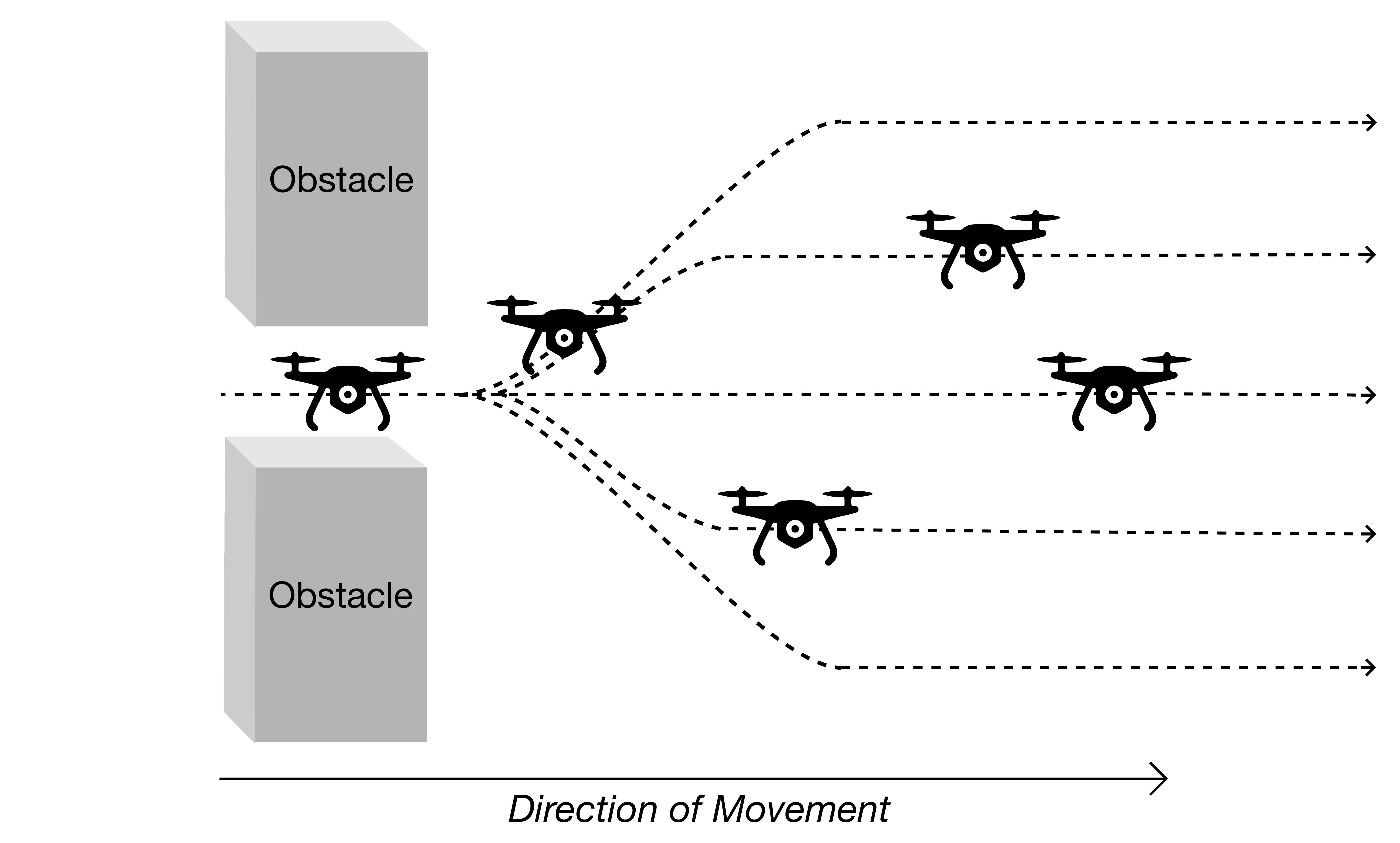}}
   \end{center}
  \caption{Illustration of the \textit{Left and Right Legs} and angles and transition phases (a) formation reshaping process while going through the obstacles, (b) Transitioning back to initial formation}
  \label{fig:navianddist}
\end{figure}

In this section we describe the proposed Dynamic Formation Reshaping Based on Point Set Registration (DFRPSR) algorithm in a Swarm of Drones. The overall strategy is to be able to navigate through the available gap between multiple obstacles while reshaping the formation in a near-optimal manner, without slowing down, and after passing through the obstacles bringing the agents back into formation dynamically while navigating towards the destination in order to minimize the overall time it takes to reach the destination. A novel top-level algorithm is developed to accomplish this, i.e., composed of partial feedback algorithms: one for realigning the agents into a queue formation for passing through the narrow openings between the obstacles and one for bringing them back into the original formation shape. Upon detection of the available gap between the obstacles in the planned route, the algorithm starts evaluating the possible reshaping solutions based on the calculated gap between the obstacles and the angle at which the available gap for safe passage lies. If the detected obstacles already lie in such a manner that the leader does not require any deviation from its original path, then the leader continues to navigate forward while its followers reshape into a queue formation by merging both \textit{Left Leg} and \textit{Right Leg} of the formation while communicating with each other to allow for an efficient integration of both legs of the swarm as shown in Figure \ref{fig:obsang}. After successfully passing the obstacles, the \textit{Turn-Back} function brings the agents back into the initial formation shape in an effective manner by utilizing the point set registration technique, as shown in Figure \ref{fig:reshape}.

\vspace{-1.0cm}

\subsection{Navigation}

\begin{algorithm}
\caption{Global Routine}\label{algo1}
\scriptsize
\begin{algorithmic}[2]
\Procedure{Navigation}{}
\State{$ObsFlag$ $\gets$ $False$;}
\State{$PSR$ $\gets$ $False$;}
\State{TShape $\gets$ Initialized based on the current state;}
\While{True}
    \State{$ObsFlag, ObsDist, ObsAng, ObsNum$ $\gets$ Obstacle\_Detection();}
    \If{$ObsFlag$}
        \State{Reshape\&Avoidance($ObsDist, ObsAng, ObsNum$);}
        \State{$ObsFlag$ $\gets$ $False$;}    
        \State{$PSR$ $\gets$ $True$;}
    \Else
        \State{Update TShape;}
        
    \EndIf
    \State{Update CShape;}
    \If{PSR}
        \State{Turn-Back(TShape);}
    \EndIf
\EndWhile
\EndProcedure
\end{algorithmic}
\end{algorithm}

The general pseudo-code of the top-level routine is given in Algorithm 1. As our initial setup, we presume that the agents are already in the defined formation shape, and the leader is dedicated and the ID assignment to the agents is completed before the mission is started. This top-level algorithm is executed by every agent locally. A Boolean variable (\textit{ObsFlag}) is initialized, whose role is to indicate the presence of an obstacle \textit{True} or absence \textit{False} (Line 2). Similarly, a Boolean variable (\textit{PSR}) is initialized (Line 3), whose role is to allow (if \textit{True}) the algorithm to call the \textit{Turn-Back} function after successful reshaping and collision avoidance or bypass the \textit{Turn-Back} function (if \textit{False}). Then based on the current position of the every agent, the target shape of the swarm \textit{TShape} is then set up, i.e., initialized, (Line 4). This shape is calculated for every agent at every time interval for the next interval where they have to navigate to. After the initialization, \textit{Obstacle\_Detection} procedure is called (Line 6). And based on the presence of an obstacle in the vicinity, the information of the characteristics of the obstacle, such as, distance, angle at which the obstacle is detected, number of obstacles are updated. If \textit{ObsFlag == True}, indicating an obstacle(s) has been detected, the Reshape\&Avoidance procedure is invoked (Lines 7-8). Once the reshaping and avoidance has been successfully performed, the \textit{ObsFlag} is reset to \textit{False} and \textit{PSR} is set to \textit{True} (Lines 9-10), indicating that the formation shape has been changed due to avoidance maneuver and now initial formation shape needs to be restored. On the other hand, if after the \textit{Obstacle\_Detection} procedure, the condition \textit{ObsFlag == False} (Line 7), indicating that no obstacle is detected, then only the \textit{TShape} is updated (Line 12). Then the current shape ($CShape$) of the swarm is updated (Line 14). Finally, if \textit{Reshape\&Avoidance} procedure was invoked earlier, i.e., \textit{PSR == True}, then \textit{Turn-Back} procedure is called for bringing the agents back into the original formation shape (Line 15-16).
\vspace{-0.3cm}

\subsection{Obstacle Detection}

The pseudo-code for the \textit{Obstacle\_Detection} procedure is given in Algorithm 2. The node, while being in this procedure, continuously keeps on scanning for the obstacles in the vicinity, and as soon as an obstacle is detected the obstacle detection flag \textit{ObsFlag} is set to \textit{True} (Lines 2-3). After the detection of the obstacle, the parameters of the detected obstacle(s) are calculated and updated (Line 4), such as, the distance of the obstacle (\textit{ObsDist}), the angle at which the obstacle is detected (\textit{ObsAng}), and the number of detected obstacles (\textit{ObsNum}).

\begin{algorithm}
\caption{Obstacle Detection}\label{algo2}
\scriptsize
\begin{algorithmic}[1]
\Procedure{Obstacle\_Detection()}{}

\If{$obs$ in $Detection\_Range$}
    \State{$ObsFlag$ $\gets$ $True$;}
    \State {$ObsNum, ObsDist, ObsAng$ $\gets$ Calculate number of obstacles, the distance and angles\par
    \hskip\algorithmicindent at which the edges lie;}
\EndIf
\EndProcedure
\end{algorithmic}
\end{algorithm}

\subsection{Formation Reshaping and Avoidance}

After the detection of obstacles, the \textit{Reshape\&Avoidance} procedure is called, pseudo-code given in Algorithm 3. This procedure is responsible for reshaping the formation, according to the situation, and successful collision avoidance. As soon as this procedure takes control, the node calculates the $Gap$ between the detected obstacles, in case of multiple detected obstacles (Lines 2-3). This algorithm works on three cases (Lines 5-18). If only one obstacle was detected, i.e., $ObsNum\not> 1$, in this case the algorithm simply calls for collision avoidance procedure, developed and presented in \cite{jdiscsic}, to perform the avoidance maneuver (Lines 7-8). If more than one obstacles were detected, i.e., $ObsNum > 1$, but the calculated $Gap$ between the obstacles is less than the defined safe distance $Safe\_dist$ indicating the gap is not wide enough for the agent to go through. In this case, the algorithm treats them as a single obstacle to perform the collision avoidance maneuver (Lines 9-10). And the third case, the condition $ObsNum > 1$ holds True, i.e., more than one obstacles detected and the gap between the obstacles more than the defined safe distance for the agents to pass through, i.e., $Gap >= Safe\_dist$ holds True (Line 11). The algorithm then checks if the $ObsAng$ is negative, as shown in Figure \ref{fig:obsang}. In this case, the leader has to move towards its right to align itself to be able to pass through the gap between the obstacles without colliding. And consequently, the \textit{Right leg}, i.e., the agents with "odd" IDs (ID= 3, 5, 7,.., n), will have to move less to be able to come into the queue formation as compared to the agents with "even" IDs (ID=2, 4, 6,.., n+1), i.e., the \textit{Left leg}, will have to move more towards their right to align properly. And therefore, the \textit{Left leg} is merged into the \textit{Right leg}, (Lines 12-13) as shown in Figure \ref{fig:reshape}. In this case, every agent follow the following protocol:

\begin{enumerate}
    \item if SELF.ID is even number
    \item make SELF.ID - 1 as temporary leader
    \item if SELF.ID is odd number
    \item make SELF.ID - 1 as temporary leader
\end{enumerate}

On the other hand, if the $ObsAng$ is positive, then \textit{Right leg} is merged into the \textit{Left leg} (Lines 14-16). And the agents follow the following protocol:

\begin{enumerate}
    \item if SELF.ID is odd number
    \item make SELF.ID - 3 as temporary leader
    \item if SELF.ID is even number
    \item make SELF.ID + 1 as temporary leader
\end{enumerate}

However, if the $ObsAng$ is neither negative or positive (Lines 17-18), meaning that the leader is already well aligned in the gap, in this case \textit{Left leg} is merged into the \textit{Right leg} as well. It is important to note here that in this case either leg can be merged into the other without any efficiency or energy difference, but instead of randomizing the routine, the routine has been hard-coded for easier function.

\begin{algorithm}
\caption{Formation Reshaping and Collision Avoidance}\label{algo3}
\scriptsize
\begin{algorithmic}[2]
\Procedure{Reshape\&Avoidance}{$ObsNum, ObstDist, ObsAng$}

\If{$ObsNum$ $>$ 1}
    \State $Gap$ $\gets$ calculated gap between the obstacles;
\EndIf
\State $STATE$ $\gets$ ($Gap$ $>$= $Safe\_dist$, $ObsNum$ $>$ 1);
\Switch{$STATE$}
\Case{(-, False)}
    \State{collision avoidance(); \Comment{single obstacle}}
    \EndCase
\Case{(False, True)}
    \State{collision avoidance(); \Comment{Treated as single obstacle}}
    \EndCase
\Case{(True, True)}
    \Comment{Multi-obstacle}
    \If{$ObsAng$ is negative}
 
        \State{Left leg merges into the right leg;}
    \Else \If{$ObsAng$ is positive}
        \State{Right leg merges into the left leg;}
    \Else
        \State{Left leg merges into the right leg;}\Comment{Priority assignment}
    \EndIf
    \EndIf
    \EndCase
\EndSwitch
\EndProcedure
\end{algorithmic}
\end{algorithm}
\vspace{-0.5cm}

\subsection{Turn-Back Function}

Once the original formation shape is distorted or reshaped, into a queue formation in our case, a procedure for bringing the swarm back into their original formation shape in a near-optimal manner is presented in this part. While bringing the swarm back to the original shape we utilize the point set registration technique \cite{8594514, DBLP:journals/corr/abs-0905-2635} that is based on the thin-plate spline \cite{854733} and is a well known methodology in data interpolation and smoothing. This technique is used as due to its non-complicated construction, complicated shapes are approximated easily using splines \cite{854733}. Here we provide and discuss for 2-dimensional formulation, where we have two sets of correlating points $X$ ({$x_i$, i = 1, 2, . . . , $n$}) and $V$ ({$v_i$, i = 1, 2, . . . , $n$}), for current shape $CShape$ and the original/target shape $TShape$ respectively. Here the coordinates of the location of a point are represented by $x_i$ = (1, $x_{ix}$, $x_{iy}$) and $v_i$ = (1, $v_{ix}$, $v_{iy}$). The mapping function $f(v_i)$ can be found by minimizing the following:

\begin{equation}
\begin{split}
    E(f) = \sum_{i=1}^{n}||x_i - f(v_i)||^2 + \\ \lambda\iint[(\frac{\partial^2f}{\partial x^2})^2 +2(\frac{\partial^2f}{\partial x\partial y})^2 + (\frac{\partial^2f}{\partial y^2})]dxdy
\end{split}
\label{etps}
\end{equation}

where $E$ is the energy function, i.e., the amount of disturbance in the formation, the integral part of the equation represents the mapping of the corresponding point sets to the correlating ones while considering the intended formation, and the scaling factor is provided by the $\lambda$. Here we set $\lambda$ to zero, as our intention is to map one point set over the other without considering the $CShape$, and therefore the closest points are mapped accordingly. Once the mapping process is performed, every node navigates towards its calculated position by utilizing the shortest path approach.
\vspace{-0.5cm}
\begin{algorithm}
\caption{Turn-Back}\label{algo4}
\scriptsize
\begin{algorithmic}[2]
\Procedure{Turn-Back}{TShape}
\While{$PSR$}
    \State{$NLoc$ $\gets$ Calculate the new coordinates for each agent;}
    \State{Temperature minimization function($NLoc$);}
    \State {Update $CShape$;}
    \If{$TShape$ == $CShape$}
        \State{$PSR$ $\gets$ False;}
    \EndIf
\EndWhile
\EndProcedure
\end{algorithmic}
\end{algorithm}
\vspace{-0.5cm}

An overview of our \textit{Turn-Back} function is given in Algorithm 4. First the next location (\textit{NLoc}) of every agent is calculated, i.e., the coordinates where every respective node should be in order to maintain the initial formation shape (Line 3). These values are then fed to the temperature minimization function for bringing the nodes to their new locations as optimally as possible (Line 4). Then the current shape (\textit{CShape}) of the swarm is updated (Line 5). Finally it is checked if the swarm has reached its intended formation shape, if the current shape matches with the target shape, the $PSR$ flag is set to False and the control is returned to the main function (Lines 6-7).

\section{Simulation Results}

The initial conditions defined for our work are as follows:

\begin{enumerate}
    \item all UAVs/agents travel with constant ground speeds
    \item the communication between UAVs is ideal, i.e., without information loss
    \item all UAVs are at the same altitude
    \item on-board localization techniques are utilized by the UAVs for obtaining their respective positions
\end{enumerate}

\begin{figure}[h]
\begin{center}
    \subfigure[\label{fig:navi}]{\includegraphics[width=0.485\linewidth]{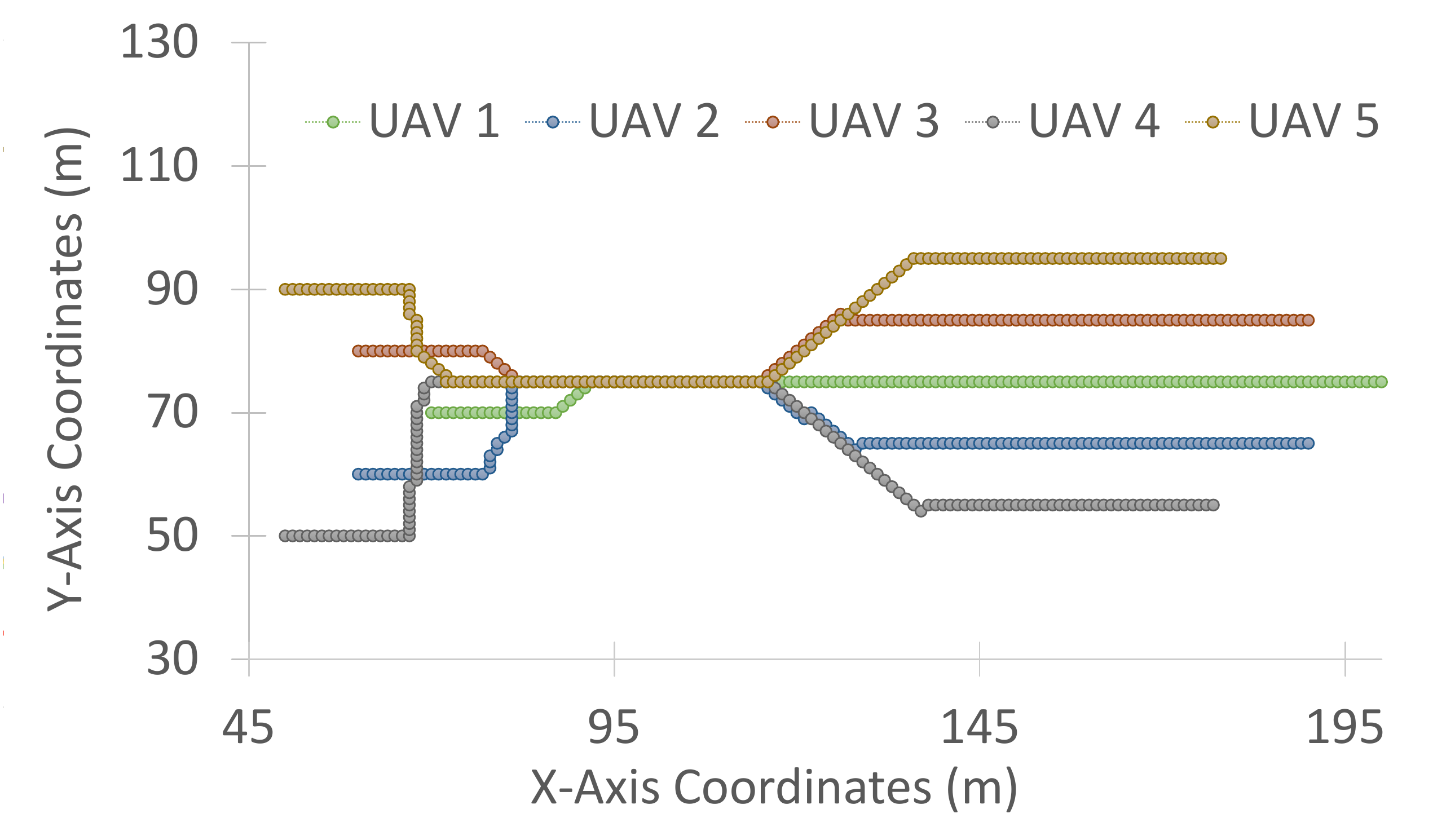}}\hspace{0.2cm}
    \subfigure[\label{fig:dist}]{\includegraphics[width=0.485\linewidth]{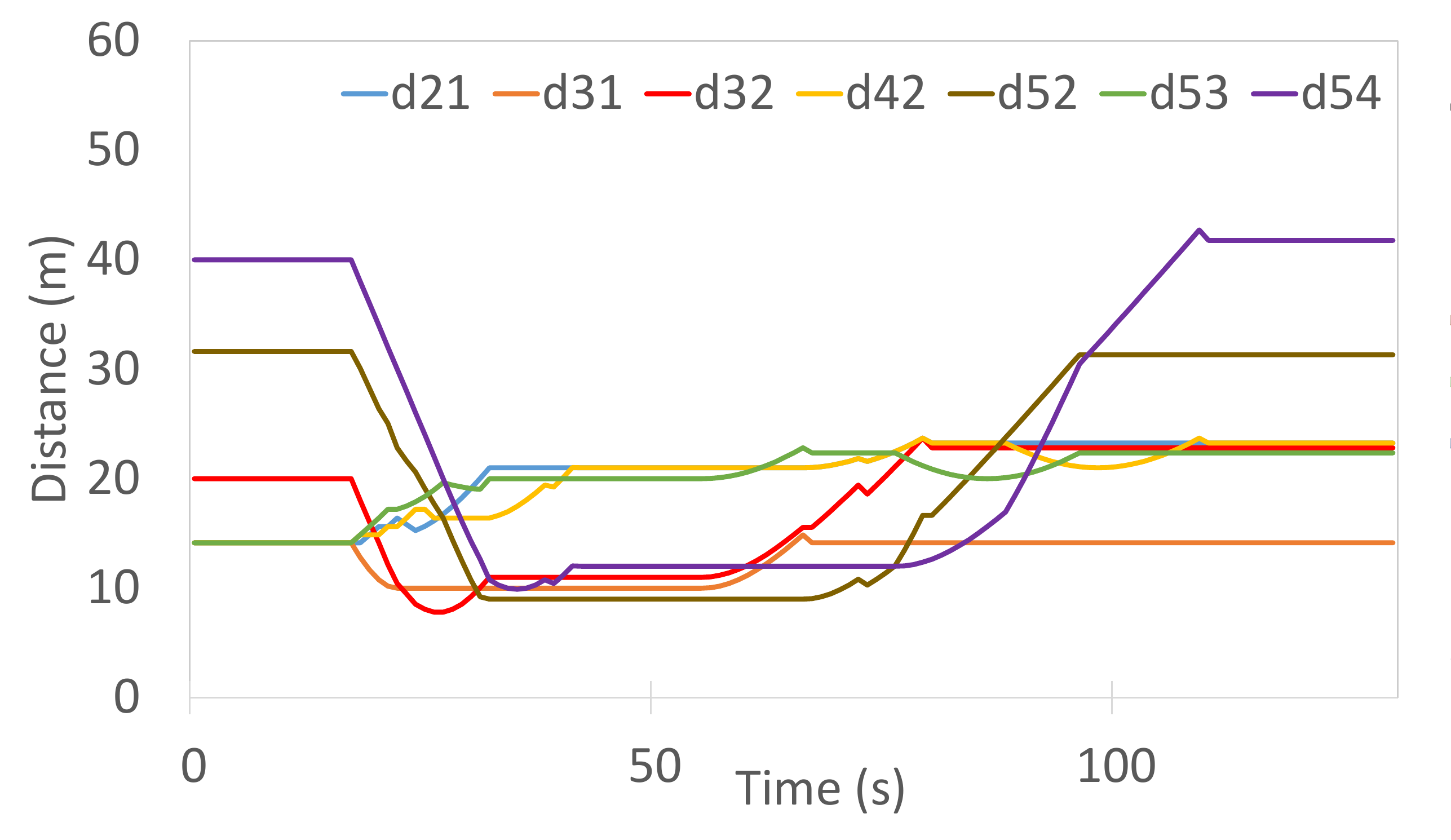}}
    \subfigure[\label{fig:distno}]{\includegraphics[width=0.485\linewidth]{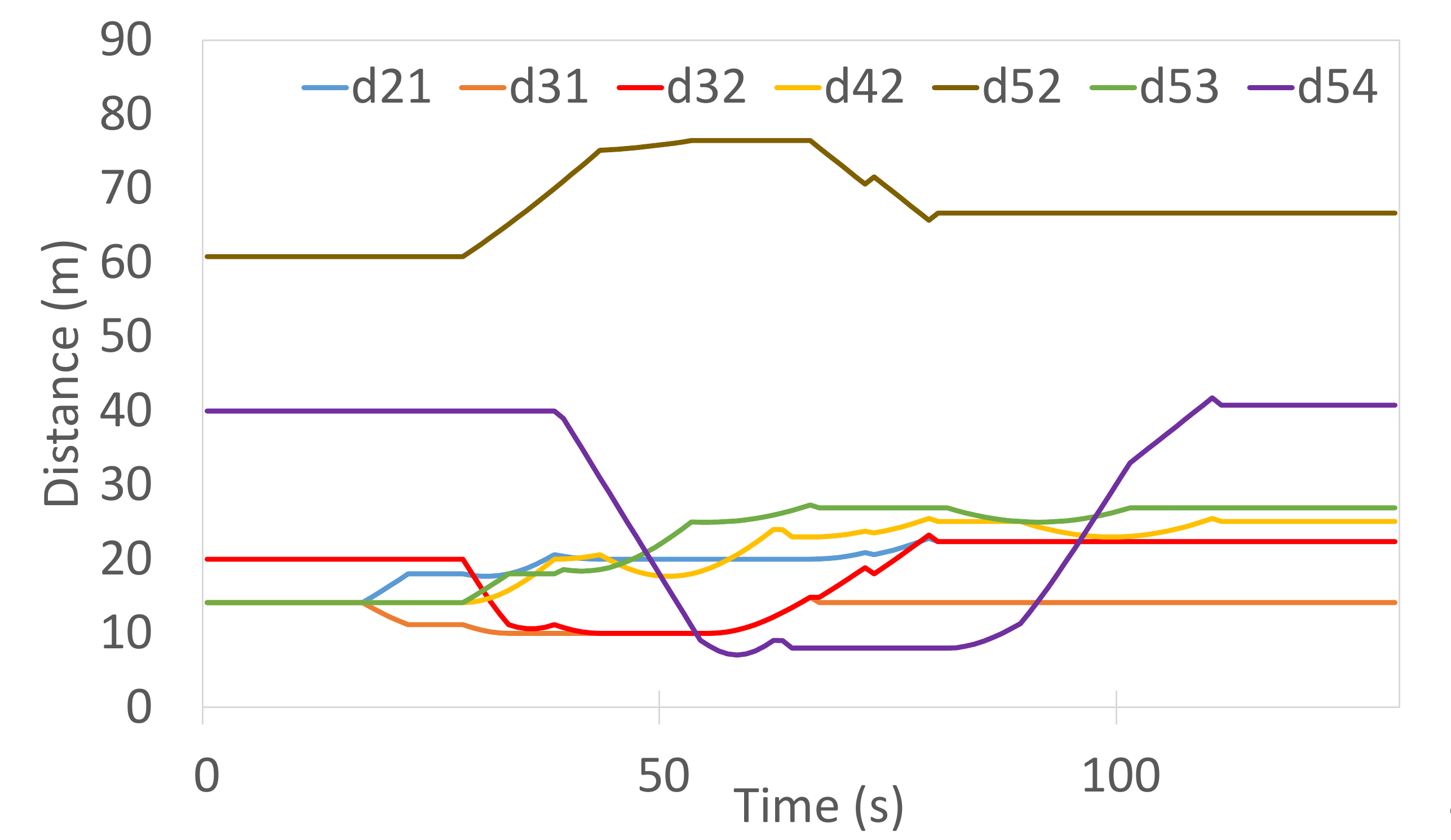}}
    
   \end{center}
   \vspace{-0.7cm}
  \caption{Simulation results (a) Overall trend of the trajectories of all UAVs. (b) Distances maintained by UAVs from their immediate leaders and temporary leaders utilizing DFRPSR algorithm. (c) Distances maintained by UAVs from their immediate leaders and temporary leaders utilizing sense\&avoid only}
  \label{fig:navianddist}
  \vspace{-0.4cm}
\end{figure}

The simulation results are shown in Figure \ref{fig:navianddist}. Figure \ref{fig:navi} shows the overall trend of the trajectories of the swarm from mission start till the destination, including the transition phases while going through the narrow opening between the obstacles and returning back to the initial formation shape. In \ref{fig:navi}, at around  $X-Axis = 85m$ the obstacle is detected by the leader, i.e., UAV 1 (green trajectory). Upon calculating the gap between them, UAV 1 realigns itself to safely pass through the obstacles and at the same time the followers are notified of the detected obstacle. As it can be seen from the figure, the followers nodes start rearranging and realigning themselves, even before the obstacle is visible to them, by merging both \textit{legs} of the swarm and following the optimal path to now follow their temporary respective leaders in a queue formation. At $X-Axis$ around 116m, the swarm coming out of the gap between the obstacles and starting the reformation process to the initial formation shape at the same time.

\begin{figure}[h]
    \centering
    \includegraphics[height= 0.5\textwidth]{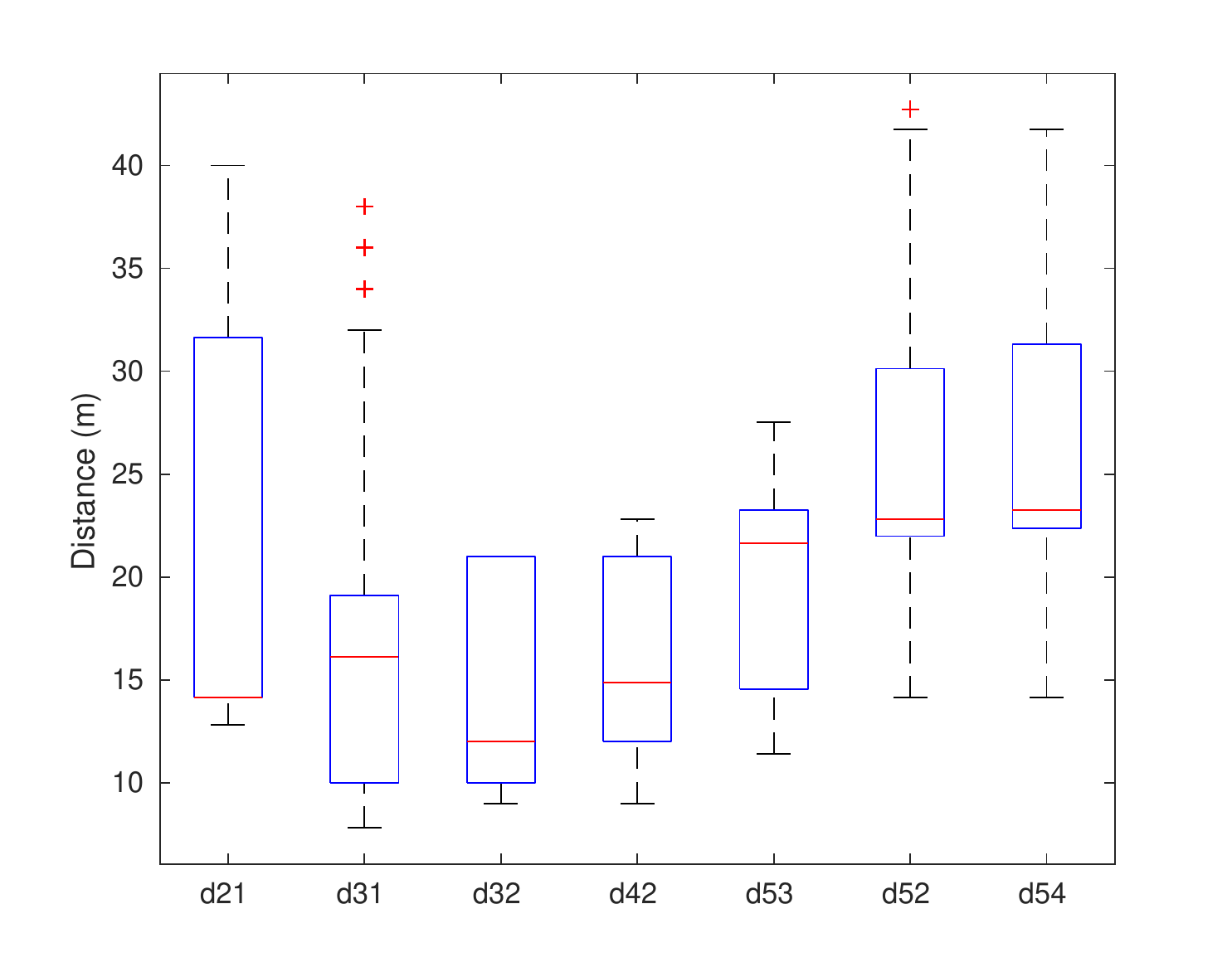}
    \vspace{0.2cm}
    \caption{Distance relationship characteristics \label{fig:dist_ave}}
\end{figure}

The distances maintained by the UAVs, with their immediate leaders and the temporary leaders (while in transition shape), through out the mission is shown in Figure \ref{fig:dist}, where distance maintained between UAV 2 and UAV 1 is represented by \textit{d21}, distance maintained between UAV 3 and UAV 1 \textit{d31}, distance maintained between UAV 3 and UAV 2 \textit{d32}, and so on. The overall distance maintenance between the nodes and their immediate leaders without utilizing the proposed DFRPSR algorithm is shown in Figure \ref{fig:distno}. In Figure \ref{fig:dist}, we show the trend of distance maintenance between the nodes and their immediate leaders only, for easier visualization and cleaner graphs. As it can be seen that as the obstacles come in the vicinity, at around $t = 17s$, the drones started reshaping, to be able to pass through the narrow passage between the obstacles, by temporarily changing their respective leaders to form a queue. From $t = 31s$ to $t = 56s$, the swarm in reformed queue shape is navigating through the gap between the obstacles. And as soon as the nodes emerge from the end of the obstacles, they start navigating to regain the initial V-shaped formation as can be seen from their trajectories and the changed inter-node distances. However, Figure \ref{fig:distno} shows that utilizing standard sense\&avoid algorithm with formation, every agent started the reshaping process based on their respective local detection. Utilizing the proposed DFRPSR approach, the swarm came back to the initial formation at $t = 97s$, whereas while local sense\&avoid methodology was applied the swarm came back to the initial formation at $t = 111s$. It shows the efficiency of the DFRPSR approach in reducing the disturbance in the formation shape more efficiently.

Figure \ref{fig:dist_ave} shows the overall trend of the minimum, maximum, and the percentile of the distances maintained by the respective nodes. Minimum distance, while navigating between UAV 1 and 2 is 12.8m, whereas the maximum distance is 40m that is due to the reshaping process and changed leaders while in transition phase, i.e., queue formation. Similarly, UAV 3 to UAV 1 the minimum distance maintained is 7.81m due to the compression between the agents while dynamic reformation process and the maximum is 38m, with three outliers. The minimum distance between UAV 4 and 2 is 9m, maximum distance was 22.82m, and the median 14.86m, that is due to the fact that UAV 2 remained the leader for UAV 4 even while in transition phase.

\section{Conclusions}

In this paper, we developed an algorithm for smooth and dynamic transformation from initial formation shape into a queue formation, by smoothly merging the agents together, to enable the swarm to efficiently pass through the gap or opening between obstacles. Upon detection of the obstacles, the leader of the swarm calculates the gap between the obstacles, aligns itself to pass through the gap, and sends that information to the followers. At the same time, the followers, start realigning and reforming the swarm by re-selecting their temporary leaders respectively. Upon successful avoidance, the nodes rearrange themselves to regain the initial formation shape as optimally as possible. The simulation results show that the proposed method works reliably and efficiently in static environments. All the agents maintained the respective defined distances between them, reshaped to maintained the minimum defined distances while being in queue formation, and expanded back to attain the original formation effectively.

The proposed approach is limited to 2-dimensional movement, with fixed altitude, in its current form and its efficiency needs to be evaluated by extending the work by introducing the third dimension. Moreover, the effectiveness of the proposed approach still needs to be tested by introducing the environmental effects such as air drag on individual drones as well as on the overall shape of the swarm. 
In the future work, we plan to further investigate the effectiveness of this approach for multi-layered formations. Another interesting analysis will be the optimization of resource management in the swarm. This will be very interesting to analyze especially when considering other environmental effects, for instance air drag on the layers, and dynamic swapping of the outer layered drones while reforming to increase the mission life.

\section*{Acknowledgement}
This work has been supported by the Academy of Finland-funded research project 314048.

\bibliographystyle{splncs03_unsrt}
\bibliography{biblio.bib}
\end{document}